\documentclass[10pt,twocolumn,letterpaper]{article}

\usepackage[pagenumbers]{cvpr} %

\usepackage{graphicx}
\usepackage{amsmath}
\usepackage{amssymb}
\usepackage{booktabs}
\usepackage{tikz}
\usepackage{listings}
\usepackage{float}
\usetikzlibrary{shapes.geometric} %

\newcommand{\thicksquarecyan}{
    \raisebox{-0.5mm}{
        \textcolor{cyan}{
            \begin{tikzpicture}
                \node[draw, thick, inner sep=0, minimum size=1em] {$$};
            \end{tikzpicture}
        }
    }
}
\newcommand{\starcyan}{
    \raisebox{-0.5mm}{
        \textcolor{cyan}{
            \begin{tikzpicture}
                \node[star, thick, star point ratio=2.25, minimum size=0.1cm, draw, scale=0.55] at (0,0) {$$};
            \end{tikzpicture}
        }
    }
}

\usepackage[pagebackref,breaklinks,colorlinks]{hyperref}

\usepackage[capitalize]{cleveref}
\crefname{section}{Sec.}{Secs.}
\Crefname{section}{Section}{Sections}
\Crefname{table}{Table}{Tables}
\crefname{table}{Tab.}{Tabs.}

\def\cvprPaperID{46
} %
\def\confName{CVPR}
\def\confYear{2024}

\newcommand{\tool}[1]{\textsc{ClickDiffusion\xspace{}}}
\usepackage{caption}

\makeatletter
\AtBeginDocument{
\let\@oldmaketitle\@maketitle
\renewcommand{\@maketitle}{\@oldmaketitle
  \includegraphics[width=\linewidth]{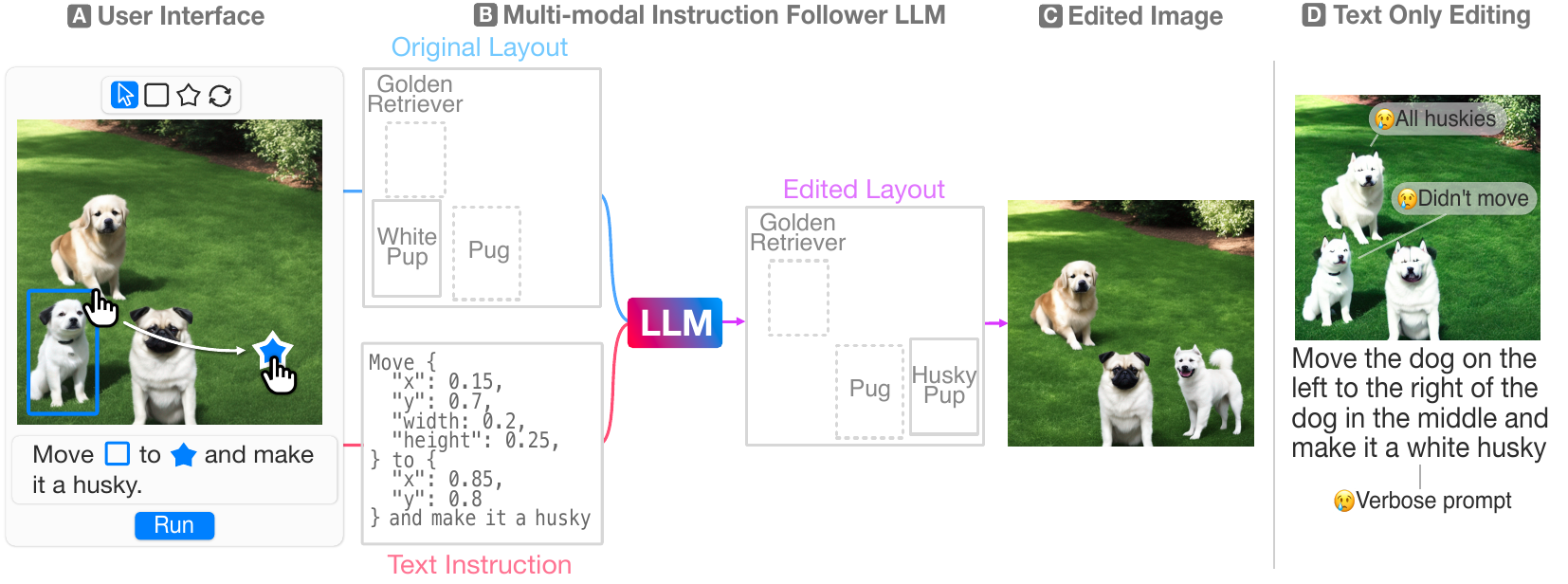}
  \captionof{figure}{
     \tool{} is an interactive system that enables users to perform fine-grained image manipulation tasks by seamlessly combining natural language and visual prompts. (A) In our example, a user can use our user interface to select a particular dog with a bounding box and a destination using a star. These locations can be referenced symbolically in a natural language instruction. (B) By serializing the original image's layout and the multi-modal instruction we can leverage an LLM to produce an edited image layout. (C) The edited layout is then fed into a layout-based image generation system to generate an edited image. (D) Our method enables moving objects and allows for much more concise prompts than text-only editing systems like \textsc{InstructPix2Pix} \cite{brooks_instructpix2pix_2023}.
  }
  \label{fig:teaser}
  \vspace{16pt}
 }
}
\makeatother 

\begin{document}

\title{ClickDiffusion: Harnessing LLMs for Interactive Precise Image Editing}

\author{Alec Helbling\\
Georgia Tech\\
{\tt\small alechelbling@gatech.edu}
\and
Seongmin Lee\\
Georgia Tech\\
{\tt\small seongmin@gatech.edu}
\and
Polo Chau\\
Georgia Tech\\
{\tt\small polo@gatech.edu}
}

\maketitle

\begin{abstract}
Recently, researchers have proposed powerful systems for generating and manipulating images using natural language instructions. However, it is difficult to precisely specify many common classes of image transformations with text alone. For example, a user may wish to change the location and breed of a particular dog in an image with several similar dogs. This task is quite difficult with natural language alone, and would require a user to write a laboriously complex prompt that both disambiguates the target dog and describes the destination. We propose \tool{}, a system for precise image manipulation and generation that combines natural language instructions with visual feedback provided by the user through a direct manipulation interface. We demonstrate that by serializing both an image and a multi-modal instruction into a textual representation it is possible to leverage LLMs to perform precise transformations of the layout and appearance of an image. Code available
\footnote{https://github.com/poloclub/ClickDiffusion}.
\end{abstract}

\vspace{-5pt}
\section{Introduction \label{section:intro}}
\vspace{-5pt}

\begin{figure*}
  \begin{center}    
    \includegraphics[width=0.95\textwidth]{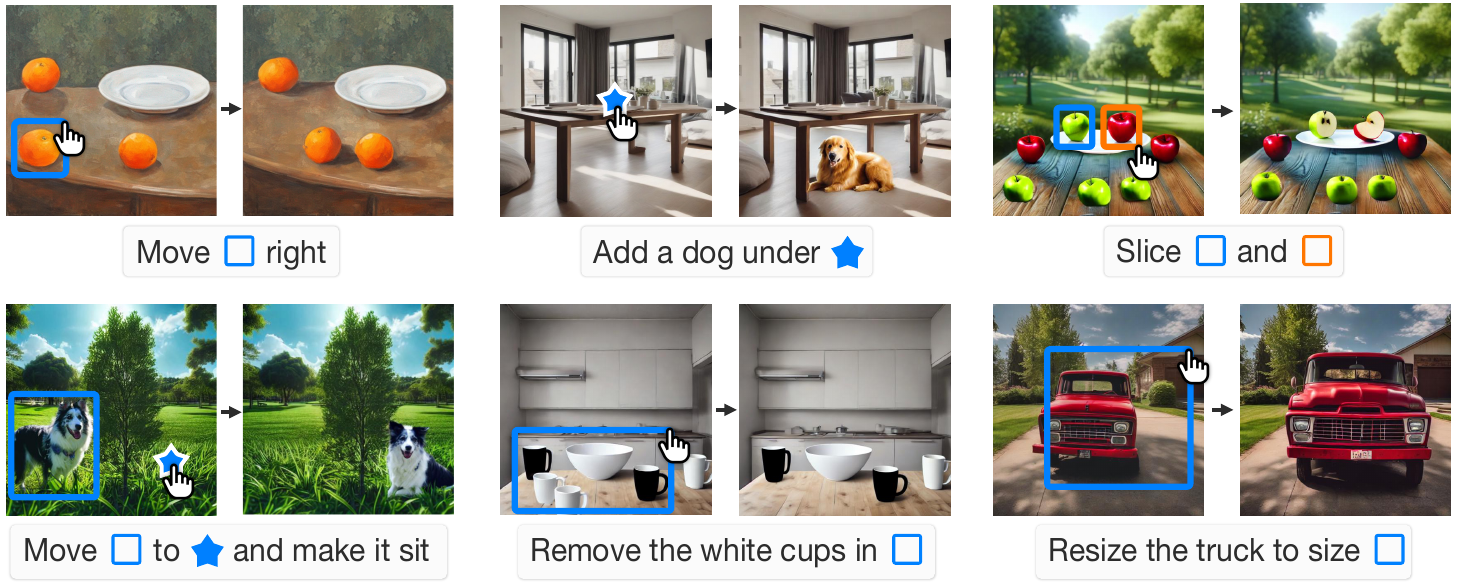}
  \end{center}
  \vspace{-0.15in}
  \caption{
  \tool{} enables users to perform precise image manipulations that are difficult to do with text alone. A user can leverage familiar direct manipulation to specify regions or objects in an image, which can be referred to in text instructions. By combining direct manipulation and natural language based editing it becomes much easier for users to perform precise edits like: moving a particular object, adding an object in a specified location, or changing the appearance of an object. }
  \label{fig:example-operations}
\end{figure*}

Recent advancements in machine learning have led to the development of powerful models capable of generating high-fidelity images from text descriptions \cite{rombach_high-resolution_2022, ramesh_zero-shot_2021, saharia_photorealistic_2022}. A similar class of models aim to empower users to edit existing images using text instructions \cite{brooks_instructpix2pix_2023, sheynin_emu_2023}. However, text alone offers users limited precision when it comes to specifying fine-grained transformations to an image. Figure \ref{fig:teaser} shows an example scenario in which a user wishes to move a particular dog in an image with multiple dogs to a precise location and change its breed. To perform this task with text alone, a user would need to expend significant effort writing laboriously complex prompts, which even then might not elicit a correct edit from an image editing system. 

Direct manipulation interfaces \cite{shneiderman1983direct, hutchins_direct_1985, shneiderman_future_1982} that leverage spatial inputs like mouse click and drag interactions offer a compelling alternative to text instructions for the task of precise manipulation of the spatial layout of an image. However, direct manipulation interfaces restrict a user to a fixed set of pre-specified operations (e.g. paint brush, polygon shapes, etc.) drastically limiting their flexibility. Motivated by the limitations of both mediums, we developed \tool{}, a system that leverages the  descriptiveness of natural language and the spatial precision of direct manipulation for the task of precise image manipulation. Our key contributions are: 

\begin{enumerate}
    \item \textbf{\tool{}, an interactive image editing system} that empowers artists and designers to make precise manipulations to images by combining natural language and direct manipulation. With \tool{} a user can specify instructions like ``move \thicksquarecyan to \starcyan and make it a golden retriever'' to select a particular object, move it to a precise location, and change its appearance (Fig. \ref{fig:teaser}). 
    \item \textbf{A novel LLM-based framework for integrating direct manipulation and text instructions for image editing. } By representing visual information like mouse interactions and bounding boxes in a textual form, we can cast the problem of image editing as one of text generation using LLMs. 
    Leveraging the few-shot generalization capabilities of LLMs allows our system to generalize to a wide range of possible transformations without the need for expensive training (Fig. \ref{fig:example-operations}). 
    To the best of our knowledge our tool is the first to leverage LLMs to integrate visual feedback with natural language instructions for image editing. 
\end{enumerate}

\smallskip
\section{Background and Related Works}

\smallskip
\noindent
\textbf{Text-based Image Editing.} Recently developed diffusion models have enabled powerful systems capable of generating high-fidelity images from text descriptions \cite{rombach_high-resolution_2022, ramesh_zero-shot_2021} and editing images from text instructions \cite{brooks_instructpix2pix_2023, hertz_prompt--prompt_2022, parmar_zero-shot_2023, sheynin_emu_2023}. 
However, many text based image editing systems are capable of making global changes to the appearance of an image \cite{brooks_instructpix2pix_2023, hertz_prompt--prompt_2022}, but struggle with tasks like moving objects or isolating changes to particular image regions. Some image-editing methods try to restrict changes to local image regions by incorporating a mask \cite{couairon2022diffedit}, similar to inpainting \cite{lugmayr2022repaint}, however these methods struggle with editing tasks that alter the layout of a scene like moving objects. In contrast, our approach allows users to both localize edits to particular objects and manipulate the spatial layout of a scene. 

\bigskip
\noindent
\textbf{Grounded Image Generation.} A key problem with text-only image generation systems is that it is quite difficult or even impossible to convey precise spatial positions of various objects in an image with text alone. This has motivated the development of grounded image generation techniques \cite{chen_training-free_2023, li_gligen_2023, lian_llm-grounded_2023} which allow users to control the positions of objects in a generated image. Another line of work aims to generate such layouts from text descriptions \cite{lian_llm-grounded_2023, feng_layoutgpt_2023, wu2023selfcorrecting} using large language models (LLMs). These works make the observation that it is possible to serialize an image layout into a JSON-like textual format, which can be readily processed by LLMs. 
Our work builds upon the observations of this line of work, demonstrating that it is possible to leverage LLMs not just to generate image layouts, but also to process a combination of visual and textual instructions for the task of image editing. 

\bigskip
\noindent
\textbf{Direct Manipulation and Natural Language Interfaces.} Direct manipulation interfaces have long been a core component of many graphical user interfaces \cite{shneiderman_1980, hutchins_direct_1985, shneiderman_future_1982}. Likewise, building flexible natural language interfaces has long been of interest in the human-computer interaction (HCI) community \cite{perrault_chapter_1988}. Several existing works attempt to demonstrate how combining direct manipulation and natural language interfaces helps overcome the limitations of both mediums \cite{cohen_natural_language_1992, cohen_synergistic_1989}. We argue that natural language is ill-suited for precise image editing tasks that require a user to disambiguate a particular object or specify a precise location. Our work attempts to solve this problem by combining text-based instruction following with direct manipulation for the task of image manipulation. Closely related to our work is that of \cite{masson2023directgpt}, where the authors demonstrate that it is possible to feed direct manipulation interactions into an LLM to perform visual manipulation tasks textual representations of visual information like SVG. However, their approach does not extend to editing real world images. In contrast, we demonstrate that it is possible to incorporate direct manipulation feedback and natural language instructions to manipulate images by leveraging diffusion models.

\bigskip
\section{Methods \label{section:methods}}

We introduce \tool{}, a system for image editing that seamlessly combines direct manipulation and natural language instructions. Our system accepts multi-modal instructions from users through a user interface, processes these instructions by serializing the multi-modal instructions into a textual form, and then leverages layout-based image generation.

\paragraph{Multi-modal Instruction Following}

Our goal is to take an input image and a multi-modal instruction and produce a transformed image that resolves the instruction. Instead of directly transforming images using a diffusion model we instead manipulate an intermediate representation of an image in the form of a spatial layout of objects specified by bounding boxes and text descriptions. This allows us to convert multi-modal instructions into a textual form that can be readily processed by LLMs.
Using our interface (Fig. \ref{fig:teaser}A) a user can specify multi-modal instructions composed of both natural language and geometric objects like bounding boxes. Suppose a user wishes to move a particular object in a complex scene. The user can select an object of interest with a bounding box \thicksquarecyan\hspace{-0.05in}, which can be represented in a textual form like  ``\texttt{\{x:~150, y:~400, width:~100, height:~100\}}'', and a destination 
\starcyan\hspace{-0.05in} represented as ``\texttt{\{x:~144, y:~132\}}''. 
By feeding an instruction into an LLM we can generate a transformed layout (Fig. \ref{fig:teaser}B), which can then be given to a layout-to-image generation system like \textsc{GLIGEN} \cite{li_gligen_2023}. Casting the problem in this way allows us to drastically simplify the problem of image manipulation, operating in image layout space rather than pixel space. Furthermore, we inherit several key advantages of LLMs like their ability to learn from just a few in-context examples and generalize to unseen vocabulary. 

Instead of performing data intensive and computationally expensive fine-tuning of an underlying language model we leverage in-context learning. In-context learning \cite{min_metaicl_2022} is a powerful technique which allows us to put examples of a task into the context window of a large-language model at inference time. We provide a textual specification of an image layout, followed by a textual multi-modal instruction. We then leverage chain-of-thought prompting by presenting the language model with intermediate questions which has been shown to improve performance on a wide variety of tasks \cite{wei_chain--thought_2023}, followed by an edited output layout. The authors hand annotated a relatively small number ($\approx 20$) of examples composed of input layouts, instructions, output layouts, and chain of thought prompts. We place multiple instances of these examples into the context window of the LLM, and at test time have an input layout and instruction. The LLM uses all of these examples to ``autocomplete'' the chain of thought and manipulation of the unseen layout and instruction.

\begin{figure*}[t!]
    \centering
    \includegraphics[width=\textwidth]{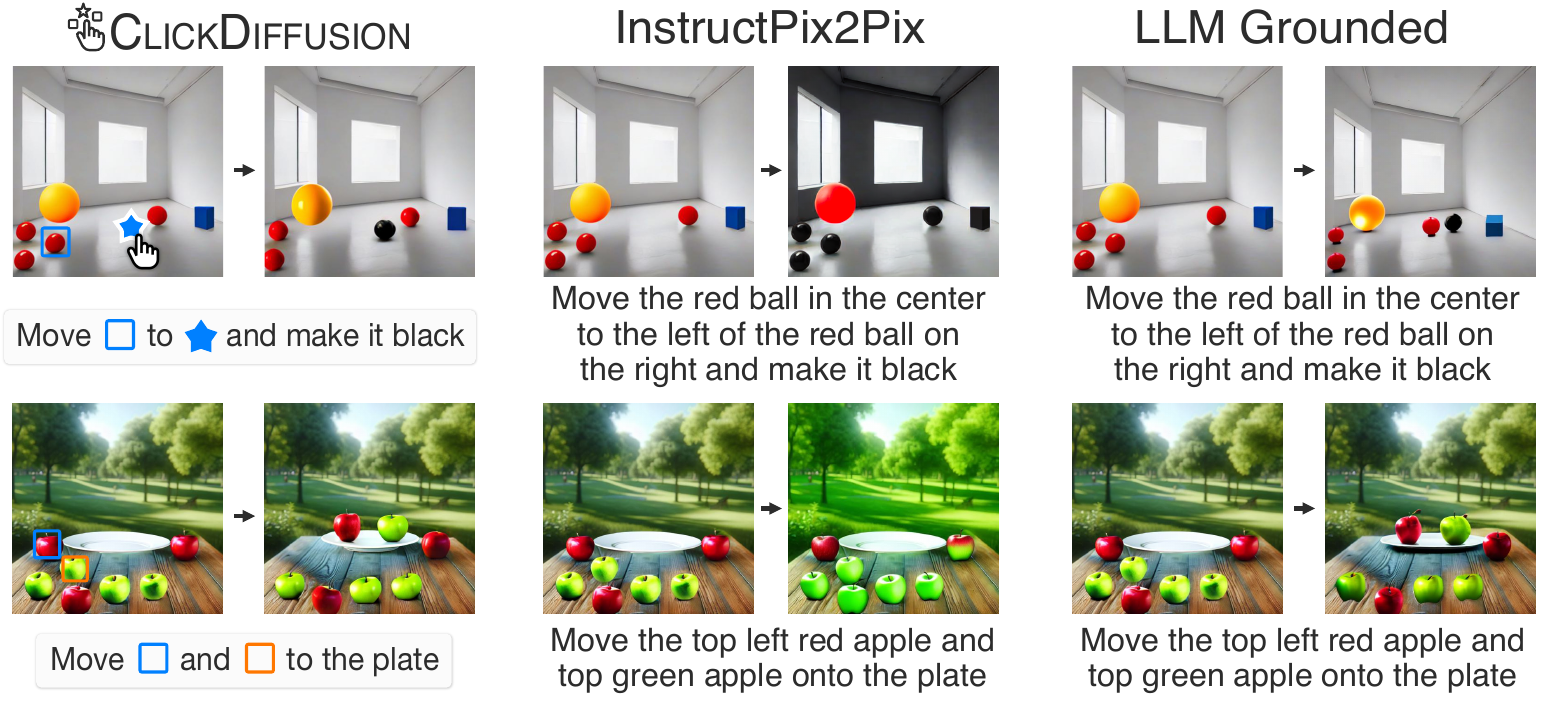}
    \caption{Our approach enables a user to disambiguate a particular object from other similar objects, move it, and change its appearance. In contrast, text-only editing approaches like \textsc{InstructPix2Pix} \cite{brooks_instructpix2pix_2023} and \textsc{LLM Grounded Diffusion} \cite{lian_llm-grounded_2023} fail to localize the manipulation to the correct object, despite requiring a much longer and more difficult to write edit instruction. }
    \label{fig:comparison}
\end{figure*}

\paragraph{User Interface} In addition to developing an LLM-based system for processing a users multi-modal instructions we aimed to develop a simple user interface that requires little familiarity with sophisticated graphic design software (Fig. \ref{fig:teaser}A). Rather than having a complex WIMP-style interface \cite{van_dam_post-wimp_1997} with many layers of dropdown menus, we simply have five tools organized in a toolbar, an interactive canvas, and a instruction input form. The tools supported by the interface are: a \textit{select} operation for selecting drawn objects, a \textit{bounding box} tool, a \textit{star} tool for specifying 2D points on the canvas, and a \textit{reload} tool for regenerating the most recent layout. Despite its simplicity, users can perform a wide array of image manipulations (see Figure \ref{fig:example-operations}). Rather than having a pre-specified tool for primitive manipulation operations like \textit{moving} an object, \textit{adding} an object, or \textit{changing the appearance of an object}, we allow user so specify these transformations using flexible language instructions. Users can leverage direct manipulation to easily specify geometric objects (e.g., bounding boxes) which are used as ``arguments'' for these natural language editing instructions. As the user draws each of these objects, they appear in the instruction text box with reference symbols as if they were words in the instruction text box.

\medskip
\subsection{Implementation Details}

To implement the user interface of \tool{} we use ReactJS \cite{react2024} with a package called Tldraw \cite{tldraw2024} to implement the interactive canvas. The front end connects to a Python Flask backend server that interfaces with the multi-modal instruction following system described in Section \ref{section:methods}. For our LLM component we leverage the GPT 3.5-Turbo API \cite{brown_language_2020} with in-context learning, although in principle our framework could be used with a number of LLMs. For layout based image generation we leverage a both GLIGEN\cite{li_gligen_2023} , which is built on top of Stable Diffusion \cite{rombach_high-resolution_2022}. However, our method is agnostic to the particular implementation of layout based generation. 

\bigskip
\section{Evaluation}

We believe it is worth highlighting scenarios where \tool{} especially shines when compared to existing text only image editing methods like \textsc{InstructPix2Pix} \cite{brooks_instructpix2pix_2023} and \textsc{LLM Grounded Diffusion} (LGD) \cite{lian_llm-grounded_2023}. For the sake of comparison, for both our approach and \cite{lian_llm-grounded_2023} we leveraged the same layout-based generation pipeline, namely a GLIGEN inpainting pipeline \cite{li_gligen_2023}. Figure \ref{fig:comparison} compares \tool{} with these two baselines on two separate editing tasks. The first task shows a user moving a particular red ball in a complex image with several red balls. With \tool{} a user can simply select the ball with a bounding box, specify its target location, and use text to describe an appearance transformation like ``make it black''. Text-only models however require much more complex queries to precisely disambiguate the correct red ball and specify the destination location. \textsc{InstructPix2Pix} fails to move the target object and fails to isolate the ``make it black'' instruction to the correct object. LGD successfully moves the correct ball, but it moves it to an incorrect location.  In the second example, a user wishes to move two selected apples onto a plate. Our approach allows a user to perform this operation successfully with a very concise instruction. However, \textsc{InstructPix2Pix} again fails to move objects, and unnecessarily changes the color of some of the apples to green. LGD is able to successfully resolve the instruction, but the instruction is much more complex and tedious to write than our system's multi-modal instruction.

\medskip
\section{Conclusion}

We introduce \tool{}, a novel system enabling user's to make precise manipulations to images by combining natural language instructions and direct manipulation. We leave a more extensive user study and quantitative evaluation of the efficacy of \tool{} for future work. 

\newpage
{\small
\bibliographystyle{ieee_fullname}
\bibliography{egbib}
}

\section{Appendix}

\begin{figure*}[t!]
    \includegraphics[width=\textwidth]{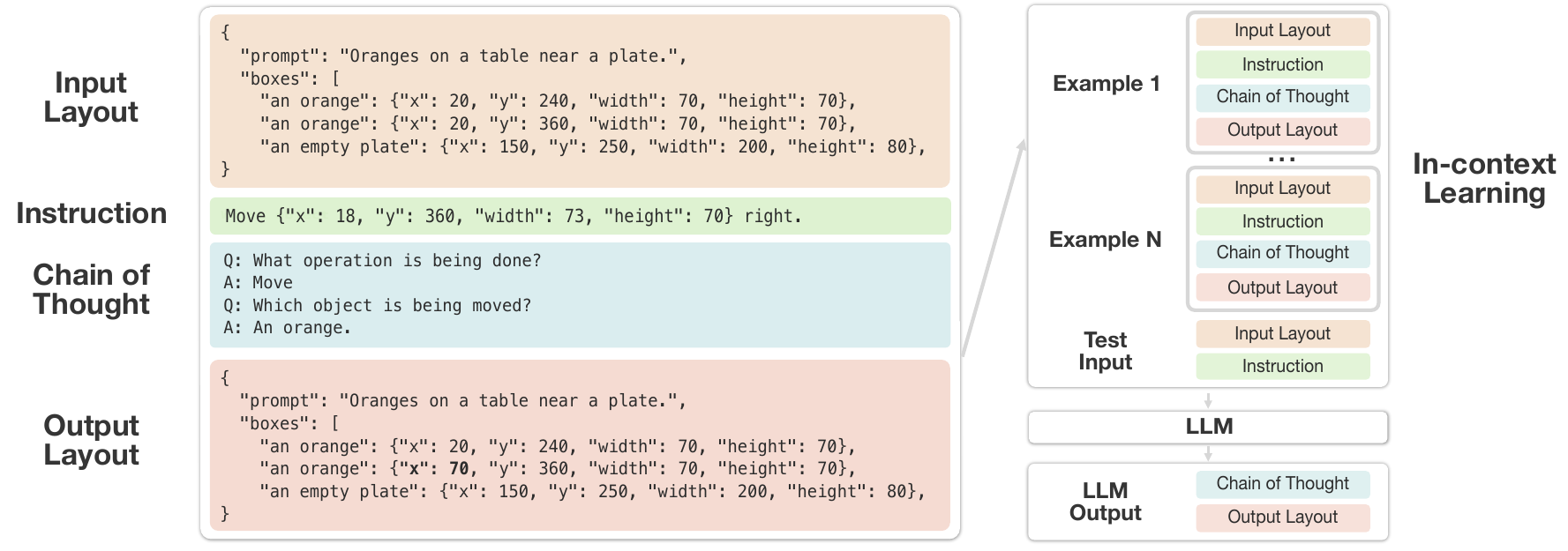}
    \caption{Our procedure for in-context learning involves placing several examples in the context of our LLM. Each example is composed of an input layout, instruction, a chain of thought, and output layout. These are placed sequentially in the context of the LLM after a preamble prompt.}
    \label{fig:in-context-learning}
\end{figure*}

\subsection{Multi-modal Prompting Through In-context Learning}  
We leverage in-context learning with a pre-trained language model for our multi-modal editing task. Figure \ref{fig:in-context-learning} shows how our in-context examples are constructed. Further, we show one such example in the text block below. 

\begin{figure*}
\captionsetup{type=verbatim} %
\begin{verbatim}
{
    "type": "text",
    "instruction": "Move the dog onto the car.",
    "chain_of_thought": "Q: Which operation is being performed? A: Move.
        Q: Which objects are being moved? A: Dog.
        Q:Which objects are not being moved? A: Car, A street
        Q: Where are they being moved to? A: Onto the car.
        Q: Does the size need to change? A: No. Is an objects apperance changing? No.",
    "input_scene_graph": {
        "prompt": "A dog standing by a car.",
        "boxes": [
            {
                "unique_id": 0,
                "name": "dog",
                "box": {"x": 0.75, "y": 0.8, "width": 0.2, "height": 0.2}
            },
            {
                "unique_id": 1,
                "name": "car",
                "box": {"x": 0.1, "y": 0.65, "width": 0.6, "height": 0.35}
            },
        ]
    },
    "output_scene_graph": {
        "prompt": "A dog standing on a car. ",
        "boxes": [
            {
                "unique_id": 1,
                "name": "car",
                "box": {"x": 0.1", "y": 0.65, "width": 0.6, "height": 0.35}
            },
            {
                "unique_id": 0,
                "name": "dog",
                "box": {"x": 0.35,  "y": 0.45, "width": 0.2, "height": 0.2}
            },
        ]
    }
}
\end{verbatim}
\end{figure*}

\end{document}